\documentclass[a4paper,twoside]{article}

\usepackage{epsfig}
\usepackage{subfigure}
\usepackage{calc}
\usepackage{amssymb}
\usepackage{amstext}
\usepackage{amsmath}
\usepackage{amsthm}
\usepackage{multicol}
\usepackage{pslatex}
\usepackage{apalike}
\usepackage{verbatim}
\usepackage{rotating}
\usepackage{graphics}
\usepackage{SCITEPRESS}     
\newcommand*\rot{\rotatebox{90}}

\newcommand\Tstrut{\rule{0pt}{2.6ex}}       

\begin{document}

\title{Joint Semantic and Motion Segmentation for dynamic scenes using Deep Convolutional Networks}

\author{\authorname{Nazrul Haque\sup{1}, Dinesh Reddy\sup{1,2} and K. Madhava Krishna\sup{1}}
\affiliation{\sup{1}International Institute of Information Technology, Hyderabad, India}
\affiliation{\sup{2}Max Planck Institute For Intelligent Systems, Tubingen, Germany}
\email{nazrul.athar@research.iiit.ac.in, dinesh.reddy@tuebingen.mpg.de, mkrishna@iiit.ac.in}
}


\abstract{Dynamic scene understanding is a challenging problem and motion segmentation plays a crucial role in solving it. Incorporating semantics and motion enhances the overall perception of the dynamic scene. For applications of outdoor robotic navigation, joint learning methods have not been extensively used for extracting spatio-temporal features or adding different priors into the formulation. The task becomes even more challenging without stereo information being incorporated. This paper proposes an approach to fuse semantic features and motion clues using CNNs, to address the problem of monocular semantic motion segmentation. We deduce semantic and motion labels by integrating optical flow as a constraint with semantic features into dilated convolution network. The pipeline consists of three main stages i.e Feature extraction, Feature amplification and Multi Scale Context Aggregation to fuse the semantics and flow features. Our joint formulation shows significant improvements in monocular motion segmentation over the state of the art methods on challenging KITTI tracking dataset.    
}
\onecolumn \maketitle \normalsize \vfill

\section{INTRODUCTION}

Visual understanding of dynamic scenes is a critical component of an autonomous outdoor navigation system. Interpreting a scene involves associating a \emph{semantic concept}, also referred to as a \emph{label} with each image pixel. These semantics can then be incorporated in a higher-level to reason about the image holistically. Traditional scene understanding approaches \cite{chen2014semantic}\cite{athanasiadis2007semantic} \cite{shotton2008semantic} have focused on extracting pixel-level semantic labels, and have demonstrated superior performance in static scenes. Motion and semantics provide complementary cues about a dynamic scene, and can be used to generate a comprehensive understanding of the scene. Some recent approaches \cite{reddy2014semantic} \cite{wedel2009detection} leverage stereo information to incorporate motion cues into the scene understanding framework.

We focus on the problem of obtaining semantic motion segmentation from monocular images. Recent success in scene understanding using convolutional neural networks, motivated us to extend existing models that perform semantic segmentation to incorporate motion cues. The success of deep neural network architectures can be attributed to the efficient learning and inference mechanisms employed. Learning involves determining a set of parameters using multiple iterations of stochastic gradient descent over randomly sampled \emph{batches} of labeled images, and inference on a target image involves only a forward pass of the image through the network.

\begin{figure}[!h]
        \centering
        \includegraphics[scale=0.28]{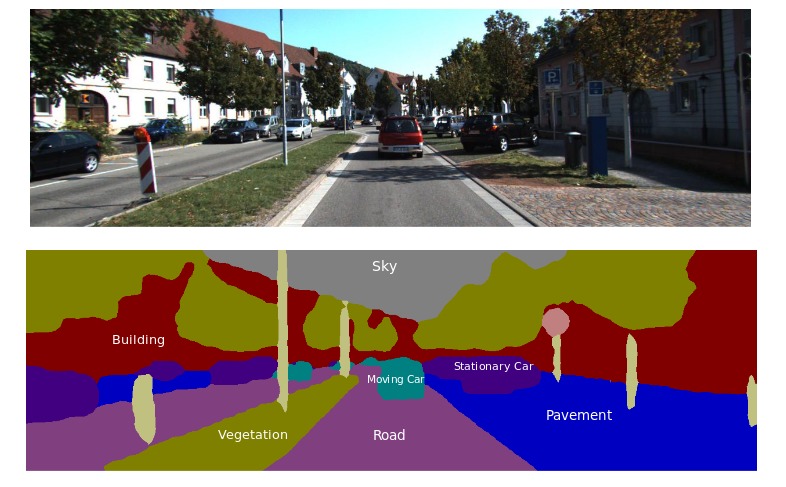}
        \caption{The eventual output of our Semantic Motion Segmentation approach. Semantic labels get prefixed with motion labels such as Moving Car and Stationary Pedestrian.(Best viewed in color) }
        \medskip
        \small
        
        \label{fig:initial}
\end{figure}

Deep learning architectures used for scene understanding incorporate semantic labels for learning scene descriptions. We aim to generate richer descriptions by \emph{prefixing} motion labels to semantics such as 'Moving Car' and 'Stationary Car', and do so in a joint framework. Currently, deep architectures model either motion \cite{fischer2015flownet} or semantics \cite{long2015fully}  in an exclusive manner. To the best of our knowledge, this is the first effort towards seamlessly integrating motion cues with deep architectures that are trained to predict only semantics. The proposed joint learning pipeline is efficient, and learning can be performed end-to-end. Fig. \ref{fig:initial} shows a sample output of the proposed framework.

In settings where images are obtained from a monocular camera, motion detection has been tackled by taking into account the optical flow between two subsequent images, which tends to fail with large camera displacements. For outdoor robotic vision, the camera displacement is unavoidable. Although, this has been tackled in \cite{tourani2016using} where motion models are generated and merged using trajectory clustering into different motion affine subspaces. The moving object proposals generated from the prior model are sparse collection of points lying on the object, resulting into a sparse motion segmentation.\cite{fragkiadaki2015learning} exploit appearance similarity to capture parts of moving objects using two stream CNN with optical flow and rank spatio-temporal segments over a video sequence by mapping clustered trajectories to the pixel tubes. In contrast, our approach performs joint optimization for pixel wise motion and semantic labels, owing to the fact that they are interrelated. An intuitive example to demonstrate the relation is that the likelihood of a moving car or moving pedestrian is more than that of a moving tree or wall. To exploit the correlation, our pipeline proposes integration of semantic and motion cues in three stages, namely, Feature extraction, Feature amplification and multi-scale context aggregation. The proposed approach is shown to be effective for motion segmentation even with a moving camera, on outdoor scenes. \newline

%


In summary, following are the key contributions of our work.
\begin{itemize}
    \item We present an end-to-end convolutional neural network architecture that performs joint learning of motion and semantic labels, from monocular images.
    \item We provide a novel method for seamless integration of motion cues with networks trained for predicting semantic labels.
    \item We present results on several sequences of the challenging KITTI benchmark and achieve results superior to the state of the art.
\end{itemize}

The remainder of the paper is organized as follows. Section 3 presents the architecture and approach used for joint learning of motion and semantic labels. In section 4, we summarize the experiments carried out, dataset used and training procedure for our joint module. We also show evaluation and comparison of our approach in section 4.3.

\begin{figure*}[t]
        \centering
        \includegraphics[scale=0.27]{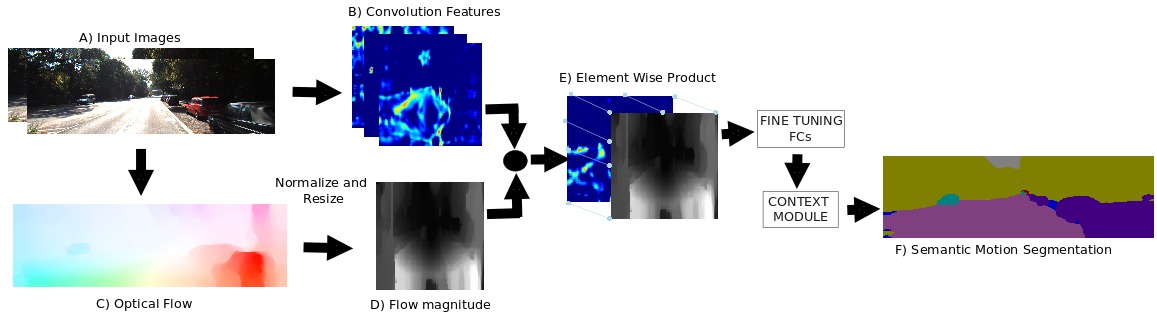}
        \caption{Illustration of the proposed approach. Images at t and t+1 are provided to the network(A). The dilated network undergoes fine tuning with addition of motion labels and the last Conv. features are extracted(B).Optical flow between the two frames(C) is scaled and resized to the size of feature maps(D). The dilated features are amplified using optical flow magnitude by element wise product(E). Further, convolution layers are freezed and fully connected layers are fine tuned. The augmented feature maps are further enhanced with end-to-end training with the Context Module, learning dependencies between object class and motion labels. The predictions obtained from the softmax layer are upsampled to give a joint label to each pixel(F).(Best viewed in color)}
        \medskip
        \small
        
        \label{fig:joint}
\end{figure*}
\section{RELATED WORK}

Fair amount of literature has been done in the field of semantic and motion understanding of scene. Traditional approaches for semantic segmentation involve extracting features from a image and use different methods to classify each pixel. Multiple works have been used to train for semantic labels \cite{fields2001probabilistic} \cite{reddy2014semantic} \cite{russell2009associative} \cite{koltun2011efficient}. However, with the emerging era of Deep Learning, there has been a large amount of literature in the field of semantic segmentation which has shown large improvements compared to the previous baselines. Approaches using deep convolutional neural networks \cite{lecun1989backpropagation} have shown to outperform most of the methods in all the basic problems of vision. The literature includes works by \cite{lin2015efficient} \cite{liu2015semantic}\cite{dai2015boxsup}, where techniques such as bounding box, Deep net followed by CRF formulation and MRFs were put to use, achieving significant results. Further, \cite{long2015fully} adapted the VGG Net model \cite{simonyan2014very} to predict pixel-to-pixel semantic labels, with fusion at pool layers for output up-sampling. Yu and Koltun\cite{yu2015multi} proposed an adaptation of VGG-16 architecture for systematic expansion of receptive fields using dilated convolutions for dense image segmentation, giving more accurate results than prior adaptations. The approach involves carrying over a global perspective without loss in resolution using repetitive deep convolutional layers.

Motion segmentation has been extensively addressed, particularly for outdoor robotic navigation. Most of the works use geometric constraints to attain significant accuracy. In the seminal work contributed by \cite{elhamifar2009sparse}, trajectory points were modeled as sparse combination of evaluated trajectories. \cite{tourani2016using} used in frame shear constraints to generate and merge affine models, achieving state of art results in sparse motion segmentation using monocular camera.  Recently many deep convolution nets have been used to learn motion labels \cite{rozantsev2014flying} \cite{fragkiadaki2015learning} \cite{tokmakov2016weakly} for motion segmentation. Although they work very well, they suffer from unavailability of large datasets or rely on stereo information, therefore proving ineffective for monocular systems. \cite{fragkiadaki2015learning} presents state of art results in the detection of per frame moving object proposals. The work emphasizes segmentation on monocular uncalibrated video sequences by a ranking heuristics and regression using a two stream network with optical flow, followed by supervoxel projection.  

Joint classification of semantic and motion labels is relatively new in the field, and much of the work has been carried out by \cite{reddy2014semantic} using dense CRF joint formulation on stereo image sequences. This however would prove ineffective for monocular situations as it heavily relies on the depth information. We draw analogy from works \cite{fischer2015flownet} \cite{simonyan2014two} \cite{karpathy2014large}  \cite{park2016combining} where two parallel streams of convolution neural networks are fused for action recognition in videos or generating optical flow. Due to unavailability of large scale datasets for semantic motion segmentation, training a neural network from scratch becomes unfeasible. However, we adapt the concept of feature amplification highlighted in \cite{park2016combining} to our problem in a joint formulation approach, resulting in an end-to-end model for semantic motion segmentation. We outperform state of art results for monocular motion segmentation using our joint model.
\vspace{-5mm}

\section{Monocular Semantic Motion Segmentation}
In this section, we present our semantic motion segmentation framework. A joint formulation is proposed for the overall learning task and is composed of three main modules, viz. features from dilated convolutions, feature amplification, and multi-scale context aggregation. We also provide an illustration of our approach in Fig. \ref{fig:joint}.

\subsection{Features from Dilated Convolutions}\label{dilatedFeatures}
To obtain semantic features, we use a neural network architecture which employs dilated convolutions, specifically engineered for dense predictions. Originally proposed in \cite{yu2015multi}, a dilated convolution operator is a traditional convolution operator modified to apply a filter at different ranges using different dilated factors.

In relation to a discrete function
    \begin{math}
    H: \mathbb{Z}^2 \rightarrow \mathbb{R}
    \end{math}
    and$ 
    \ q: \Omega_{s} \rightarrow \mathbb{R}
    $
    , a discrete filter with size
    $(2s + 1)^2$, where $\Omega_{s}  = [-s, s]^2 \cap \mathbb{Z}^2$,
    the convolution operator is defined as: \\
    \begin{equation}
    (H *_{d} q)(\boldmath{c}) = \sum_{r+dt=c}{H(r)q(t)}
    \end{equation}
    
    where, \begin{math} d \end{math} is the dilation factor.
    Such an operator \begin{math} *_{d} \end{math} is referred to as \begin{math}d \end{math}-dilated convolution. The operator can be intuitively understood as follows. Given a 1D signal \begin{math} f \end{math} and a kernel \begin{math} q \end{math}, with dilated convolution the kernel touches the signal at every \begin{math} d^{th} \end{math} entry.

Expansion of receptive fields in existing pooled architectures leads to an ungainly increase in parameters to the same extent. The architecture proposed by Fisher et. al. \cite{yu2015multi} is inspired from the fact that dilated convolutions sustain exponential expansion of the effective receptive field without loss in coverage area. While pooling architectures leads to loss in resolution, the dilated architecture enables initialization with the same parameters and producing higher resolution output.
    \begin{figure}[h]
        \centering
        \includegraphics[scale=0.37]{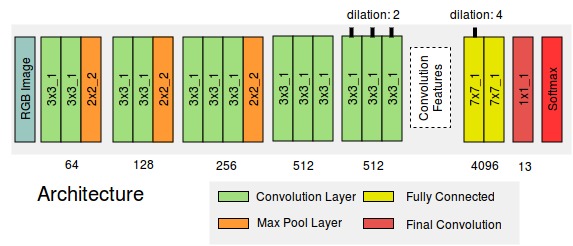}
        \caption{Network Architecture - $w\times h$\_s: Layer with kernels of width $w$, height $h$, and stride $s$. The dilated factor in layers, if any, is shown on the top of each layer. Number of channels in the outputs from each layer is depicted below each layer. For instance, the fully connected has 4096 channels in its output block.}
        \medskip
        \small
        
        \label{fig:basicnet}
    \end{figure}
\subsubsection{Network Architecture}  
Our network architecture is primarily adapted from the VGG-16 framework proposed by \cite{simonyan2014very}, with modifications applied from the work by Fisher on dilated convolutions.
The VGG-16 architecture incorporates a stack of convolutions, followed by three fully-connected layers. This was tailored for dense predictions by Long et al. \cite{long2015fully}. The architecture proposed by Long includes two major shifts. First, the inner product layers are converted to convolutions. This overcomes the restriction on the size of the input image owing to the fact that the architecture does not contain any inner product layers. Second, an upsampling layer is introduced, which brings back the spatial resolution of the output through a learned operation. The upsampling operation is carried out at different intermediate layers and are fused to obtain dense predictions. This allows the architecture to predict finer details with global or high level information in place.

We adapt the fully convolutional network of Long. and integrate modifications proposed by \cite{yu2015multi}.
Our network architecture is shown in Fig. \ref{fig:basicnet}. The last two pooling layers in the VGG-16 architecture \cite{simonyan2014very} were removed. Furthermore, for each of the removed layers, the following convolution layers are replaced with a dilation factor of \begin{math} 2 \end{math}. This enables the network to generate high resolution features with the same initialization parameters.

\subsubsection{Network Initialization}
We present a novel method for initialization of the ConvNet for obtaining convolution features. We use the model by \cite{yu2015multi}, pre-trained for semantic labels. For training with joint semantic and motion labels, we modify the final convolution layer and change the number of outputs to (C+M), where C is the number of semantic labels predicted by the dilated ConvNet and M is the number of motion labels. For instance, M can be 2, with two labels being moving car and moving Pedestrians in an outdoor scene. Furthermore, we copy the weights from the pretrained dilated ConvNet to the modified network for all layers except the final layer. For the final convolution layer, we copy the weights in the given fashion:


        \begin{align*}
            weights['final'][i] \rightarrow data[1:C, :, :, :] \\
           = weights_p['final'][i] &\rightarrow data
        \end{align*}
    
            where, $i \in\{0,1\}$,  1 for weights and 0 for bias,   
               $weights_p$ is the pre-trained weights array of the dilated network for semantic features and $weights$ is the weights array of the modified network. Weights for M motion labels in the final convolution layer are initialized using Xavier initialization \cite{glorot2010understanding}.
        \\\\
We propose that the initialization scheme works well for training with fairly small annotated datasets. The proposed initialization subjugates the limitation of unavailability of large scale annotated dataset to perform training for joint semantic and motion labels. The network is trained on our annotated dataset with the given initialization. Furthermore, the 'Convolution features'(see Fig.\ref{fig:basicnet}) from the network are extracted for joint learning with flow features. Also, the joint labels obtained from the softmax layer forms our \textit{Baseline} results for future comparisons.
    
\subsection{Feature Amplification}
We leverage optical flow for learning motion cues in an image. Conventionally, training two stream networks is found useful to the task where one is focused on learning semantic features using RGB image input, while the other is tasked for learning motion cues. The features from the two streams are fused at an intermediate layer for joint learning. However, unavailability of a large annotated dataset with joint semantic and motion labels is a major bottleneck for learning with two stream architectures. Akin to the ideas proposed in \cite{park2016combining}, we present an approach for learning relationship between semantic and motion class of an object. The method proposed in \cite{park2016combining} is used primarily for action recognition tasks. Features from the last convolution layer in a Convolutional Network tasked for learning semantic features is amplified using optical flow magnitude to identify the moving parts in an image before the fully connected layers are evaluated. However, we extend the underlying idea for the task of semantic motion segmentation. An intuitive reasoning behind such an adaptation is the similarity in recognition of motion cues and integration with semantic features in both the problems.

\begin{figure}[h]
        \centering
        \includegraphics[scale=0.32]{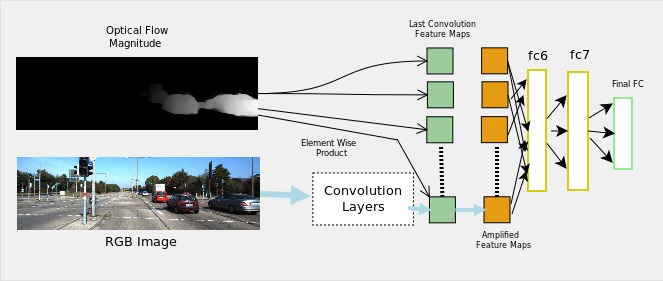}
        \caption{Fine tuning fully connected layers using amplified features as described in \cite{park2016combining}. Amplified features are obtained by taking element wise product of scaled optical flow magnitude with the feature maps obtained from last convolution layer.  }
        \medskip
        \small
        
        \label{fig:flow}
    \end{figure}

 We propose a method to augment feature maps obtained from the last convolution layer of our dilated network (see Fig \ref{fig:basicnet}), for incorporating motion cues. Optical flow is generated between the consecutive frames at t and t+1. Next, we compute Euclidean norm of the flow vector and normalize the magnitudes in the range \begin{math}
1\end{math}-\begin{math}2\end{math}. 
With the flow information in hand, we quantize the scaled magnitudes and subsequently convert it to grayscale image. The image is further resized to the size of feature maps of the last convolution layer obtained from our spatial network. Given a 900x900 RGB image as input, our dilated network outputs feature maps of dimension 512x90x90. Hence, flow image is resized to 90x90 dimension. Thereafter, element wise product is performed between the flow image and each feature map in the stack. The intuition of scaling the magnitudes from 1 rather than 0 is to not zero out the feature values obtained from the spatial network, which is equally important. Further, we \textit{freeze} the convolution layers of the network and fine tune the fully connected layers with the amplified feature maps as input to the fully connected layers. The amplification process is visualized in Fig. \ref{fig:flow}. The semantic features are enhanced with motion cues, as a consequence of feature amplification. We benefit with the amplification due to incorporated temporal consistency with optical flow and difference in flow magnitude between moving objects and it's surroundings. Also, object boundaries are retained due to amplification over baseline semantic motion features, thereby handling disorientation in optical flow boundaries. Label probabilities obtained after fine tuning from the softmax layer are up-sampled to obtain dense predictions with joint labels. Image predictions obtained forms our \textit{Joint} results for evaluations.



\subsection{Multi-Scale Context Aggregation} \label{contextModule}
We use the context module introduced by Fisher \cite{yu2015multi} for enhancement of the amplified features. The architecture was proposed as an extension to existing CNN architectures for overall increase in accuracy for dense predictions.  
The module improves upon the feature maps, by successive dilated convolutions, supporting exponential expansion of receptive field, without losing resolution. This is effectuated by continuous increase in dilation with increasing layer depth. The architecture consists of $7$ convolution layers. The layers are dilated with factors - 1, 1, 2, 4, 8, 16 and 1, and each of these layers apply 3x3 convolutions with the specified dilation factors. The module aggregates contextual information at multiple scales and outputs feature maps of the same size as that of input by padding the intermediate layers.

The weights in this module are initialized with a form of identity initialization, commonly used for recurrent networks. In mathematical terms:
        \begin{equation}
            q^j(\textbf{t},i) = 1_{\textbf{t}=0}1_{i=j}
        \end{equation}
        where i and j are index of input and output feature maps respectively.
    The identity initialization of such a form, initiates filters which can relay the inputs to the next layer.

We learn the parameters for the context module with our amplified feature maps as input to the module. Fully connected and softmax layers from our network is appended to the module. We obtain joint label predictions from the softmax layer.                   
    


\section{Experiments and Evaluation}

Experiments were carried out with pre-existing architectures, adapted to our problem. Concept of two stream architectures have been recently used in the field of action recognition, where spatial and temporal nets are combined at the fully connected layer. We tailor the architecture to our problem. Two VGG-16\cite{simonyan2014very} networks with image and optical flow as input to the respective networks were trained on our annotated dataset. The weights for both the streams were initialized with VGG models pre-trained for semantic segmentation task. We also inspect and implement Flow net \cite{fischer2015flownet} to our problem, which has shown to outperform state of art in learning optical flow. The network was initialized with pretrained FlowNet-C weights and trained on our annotated dataset with inputs as image at t and t+1 respectively for the two streams. However, both the formulations  did not work well in combining motion cues with semantic information in hand. This is attributed to the failure of CNNs in learning and extracting useful features with smaller datasets. Collection of large scale scene datasets with joint semantic and motion labels is very expensive. In contrast, our joint learning approach reduces the burden of learning motion features from scratch with large labeled datasets and proves effective with fairly smaller annotated datasets. \\
  In this section, we describe the details of ConvNet training and evaluations on KITTI tracking dataset.

\subsection{Dataset} \label{dataset}
We have used renowned \textit{KITTI} dataset \cite{Geiger2012CVPR}, for evaluation of our approach. The dataset contains over 40,000 images taken by a camera mounted on a driving car through European Roads. The driving sequences contain images from residential and urban scenes posing it as a challenging dataset. The dataset was chosen to showcase proficiency of our approach with multiple moving cars for outdoor scenes, which is uncommon in other datasets. 40 images were chosen from five sequences each, giving 200 images for training. Each of the images were manually annotated with 13 labels. To be specific, the labels given were \textit{Building, Vegetation, Sky, Car, Sign, Road, Pedestrian, Fence, Pole, Sidewalk, Cyclist} and \textit{Moving Car, Moving Pedestrian} for objects in motion. For testing, 60 images from KITTI tracking sequences were chosen as validation set and annotated with the given label spectrum. For validation set, we use challenging sequences with multiple moving cars and ensure no overlap between train and validation deck.  We have used DeepFlow\cite{weinzaepfel2013deepflow} for dense optical flow computation, known for its state of art results for KITTI benchmark dataset. We plan to release the code, trained models and dataset with joint labels to encourage future work in the field.  


\subsection{Learning}
In this section we describe the training procedure for our proposed approach.
Our implementation is based on publicly available Caffe\cite{jia2014caffe} framework. First, we describe the input to the data channel in the network. This applies to all modules in our proposed method. Input image resolution is 1242 x 375, obtained from KITTI tracking dataset. Images are padded using reflection padding and 900x900 random crops are sampled. It then undergoes randomized horizontal flipping. Further, each input batch contains crops from randomly selected images from the training dataset. This shapes the input to the module.

\textbf{Training}: Training is performed in three stages. At first, the network architecture (see Fig. \ref{fig:basicnet}) is fine tuned with motion labels added, to obtain convolution features and weights initialization for joint training with flow features. Learning rate and momentum was set to $10^{-4}$  and 0.9, respectively. Training was carried out for 10,000 iterations with batch size 1, using stochastic gradient descent. The dense predictions obtained from the module forms our \textit{baseline} for further comparisons.
We use these learned weights to train the joint model with augmented feature maps as input. Optical flow magnitude is computed between the frame at t and t+1. Flow image is padded and cropped to 900x900, with respect to the RGB crop. Furthermore, the convolution layers are \textit{freezed} and the network is trained with the amplified feature maps as input to the fully connected layers. Training was carried out for 10,000 iterations. Other parameters stay the same.       

Then, the context model is plugged into the architecture and end-to-end training is performed for 20,000 iterations with batch size 1. Learning rate and momentum is set to $10^{-5}$ and 0.99, respectively. We
refer joint label predictions obtained from the softmax layer of this model as \textit{Joint+Context}.
\begin{figure}[h!]
        \begin{center}
        
        \begin{tabular}{c c}
        \begin{sideways}\bf \centering  \ {\small +Context}\ \ \  \small Joint\ \ \  Baseline\ \ \ \ \ \small INPUT \end{sideways}&\includegraphics[width=50mm]{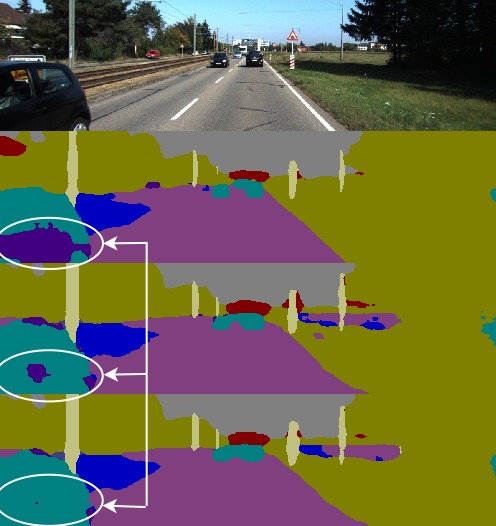}
        \\
        & \multicolumn{1}{c}{\includegraphics[width=50mm]{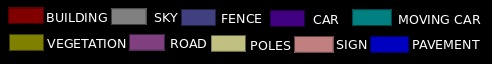}}
        \end{tabular}
        \end{center}                
        
        \caption{Figure outlining the labels from each stage of our end-to-end module. Image is taken from our KITTI tracking test dataset. The baseline predictions outputs wrong labeling to moving car patches and the motion labels of the car[Cyan] improve significantly using our joint model.(Best viewed in color) }
        \label{fig:baselineimprovement}

\end{figure}
\begin{figure*}[t!]
\begin{center}
\begin{tabular}{c c c c c}
\begin{sideways}\bf \centering\ \ \ \ \ RGB\end{sideways}&\includegraphics[width=65mm]{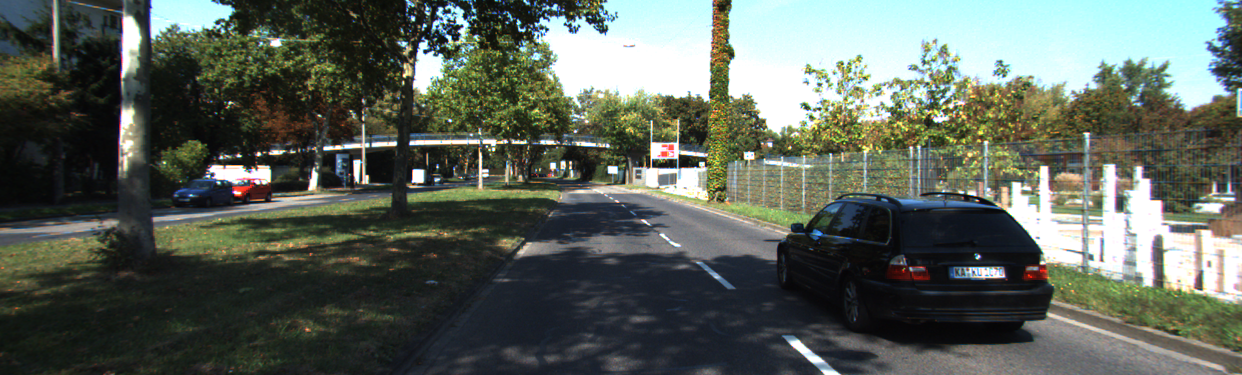} & \includegraphics[width=65mm]{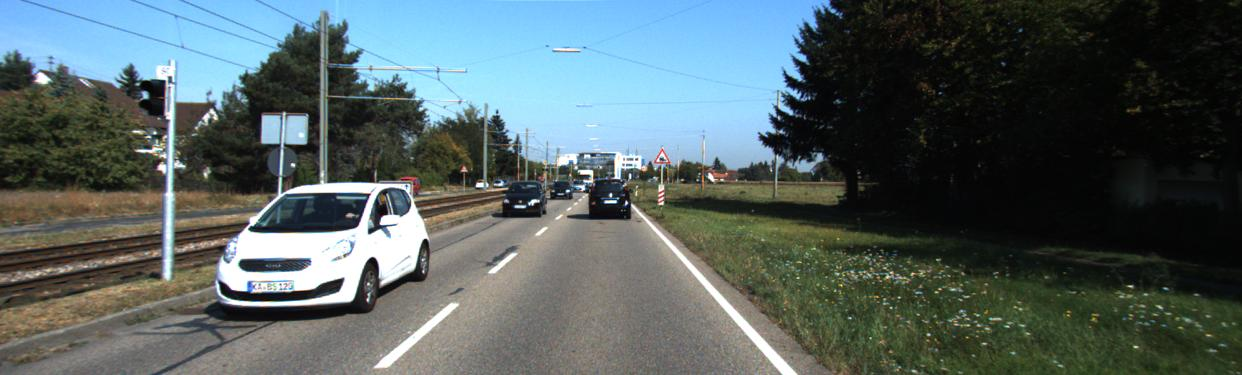} \\

\begin{sideways}\bf \centering \ \ \  \ GT - M\end{sideways}&\includegraphics[width=65mm]{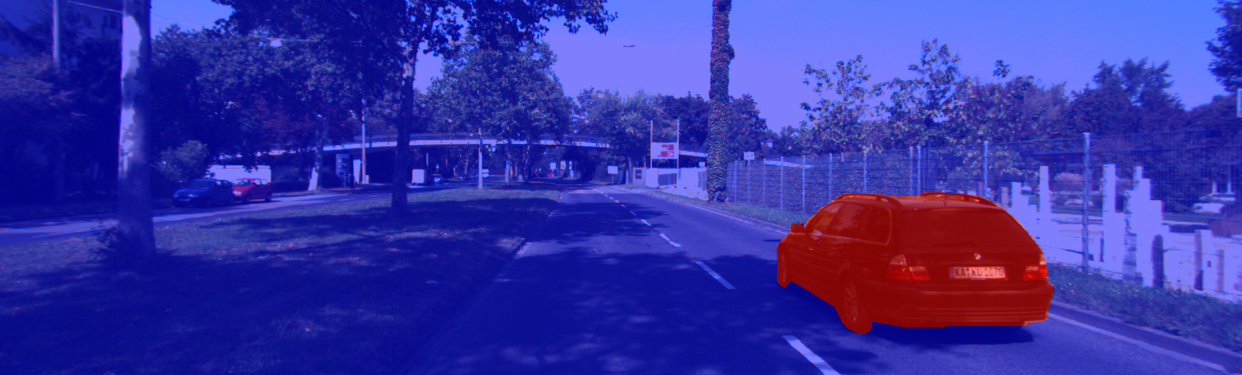} & \includegraphics[width=65mm]{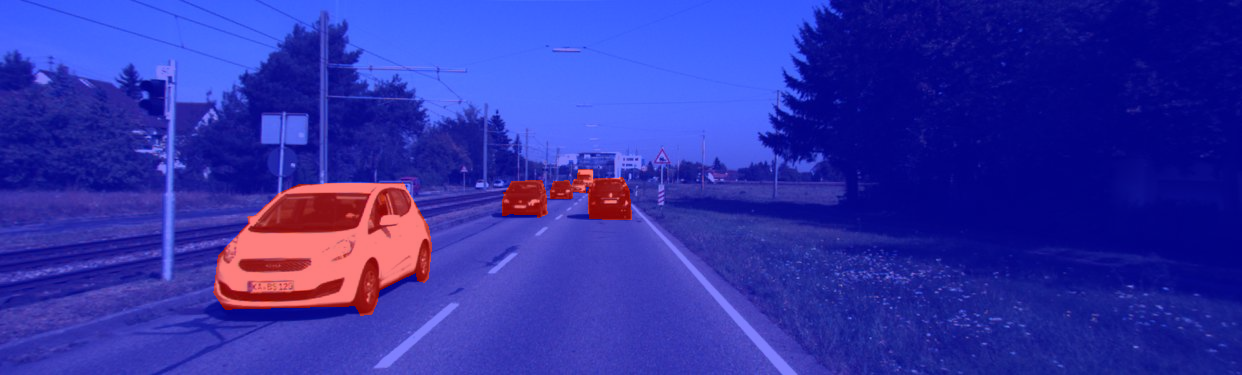} \\

\begin{sideways}\bf \centering \ STMOP - M\end{sideways}&\includegraphics[width=65mm]{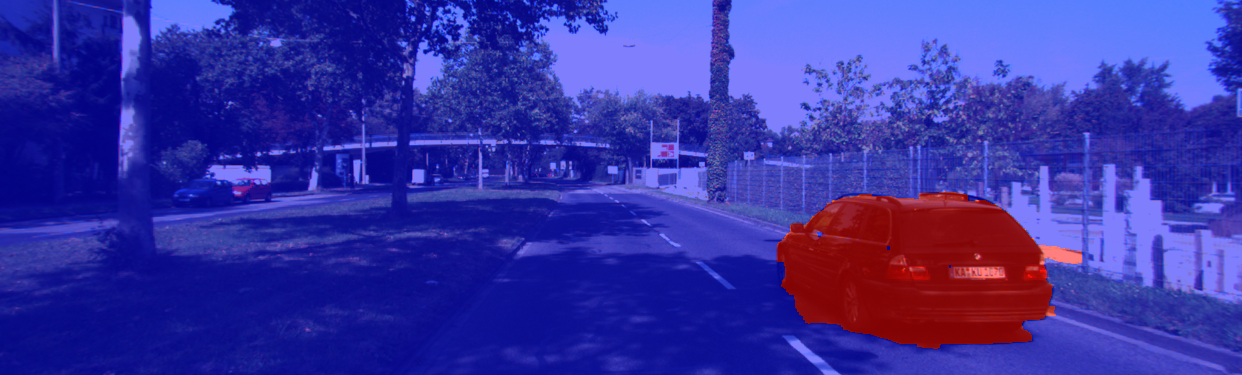} & \includegraphics[width=65mm]{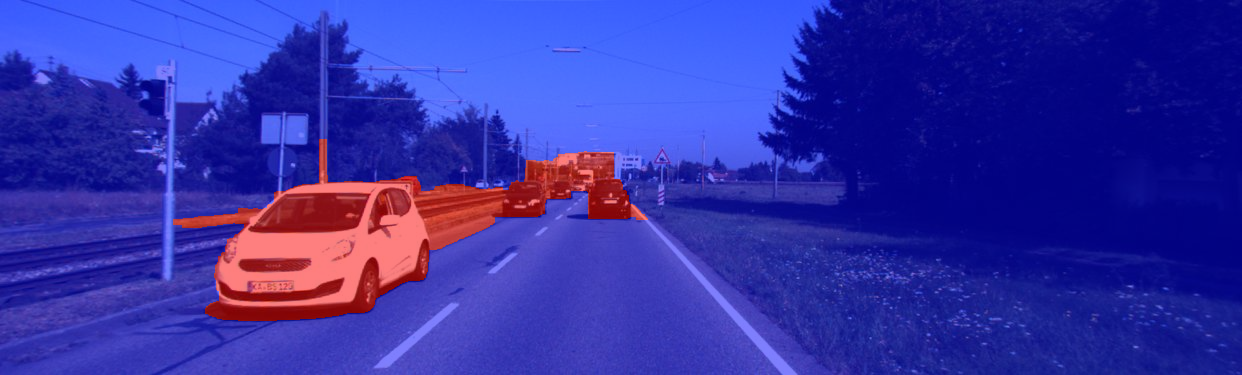} \\

\begin{sideways}\bf \centering \ \ OURS - M\end{sideways}&\includegraphics[width=65mm]{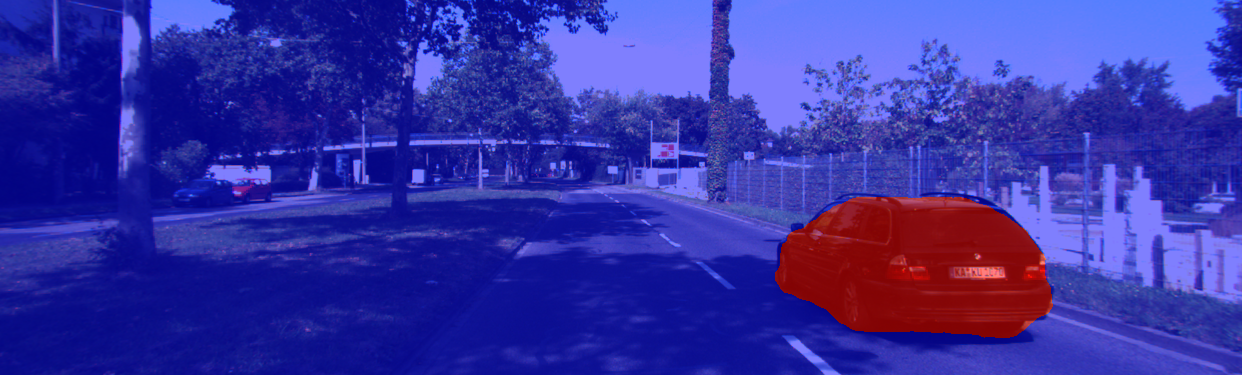} & \includegraphics[width=65mm]{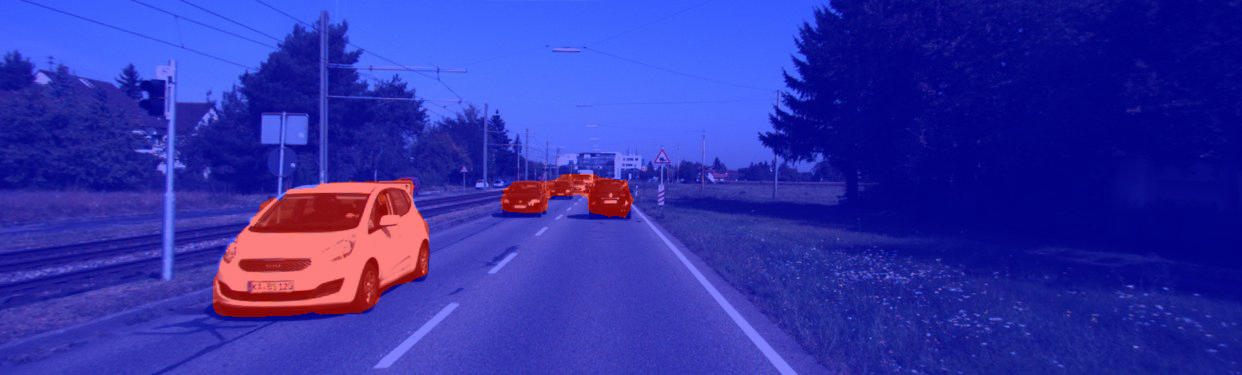} \\

\end{tabular}

\end{center}

 \caption{Qualitative evaluation - Motion segmentation on our KITTI test dataset. On the left, the images, consist of single moving car. On the contrary, we have multiple cars in the image, on the right. Blue pixels represent stationary and red pixels depict motion. We compare our approach with STMOP-M moving object proposals \cite{fragkiadaki2015learning}. In the figure, GT - M is ground truth motion annotation, STMOP - M is output from \cite{fragkiadaki2015learning} and OURS - M  is the motion segmentation obtained from the proposed approach. In contrast to STMOP-M where over-segmentation and False Positive cases are observed on the roads and fence, our proposed approach yields better segmentation and motion boundaries with cars in motion.   (Best viewed in color)}
\label{fig:qualitative3} 
 \end{figure*}
 \begin{figure*}[t!]
\begin{center}
\begin{tabular}{c c c c}
& \bf Sequence 1 & \bf Sequence 2 & \bf Sequence 3  \\
\begin{sideways}\bf \centering  \ \ \ RGB\end{sideways} & \includegraphics[width=47mm]{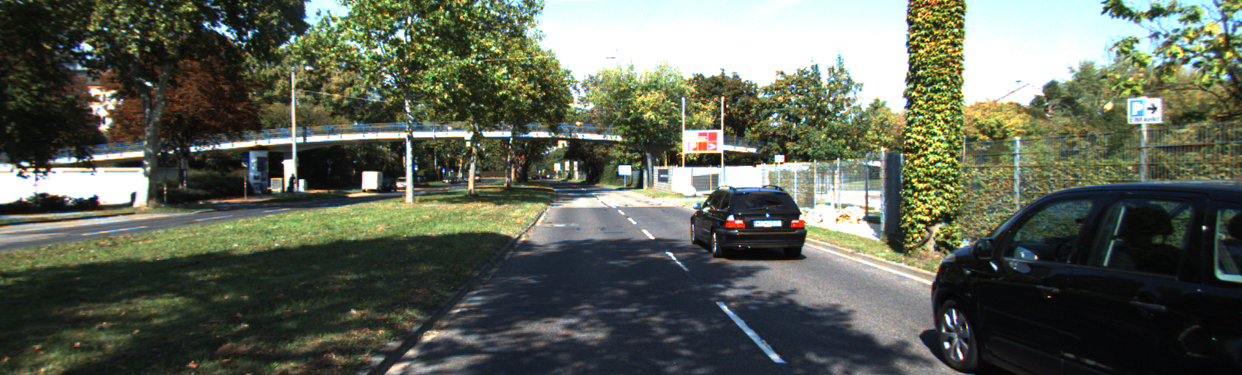} & \includegraphics[width=47mm]{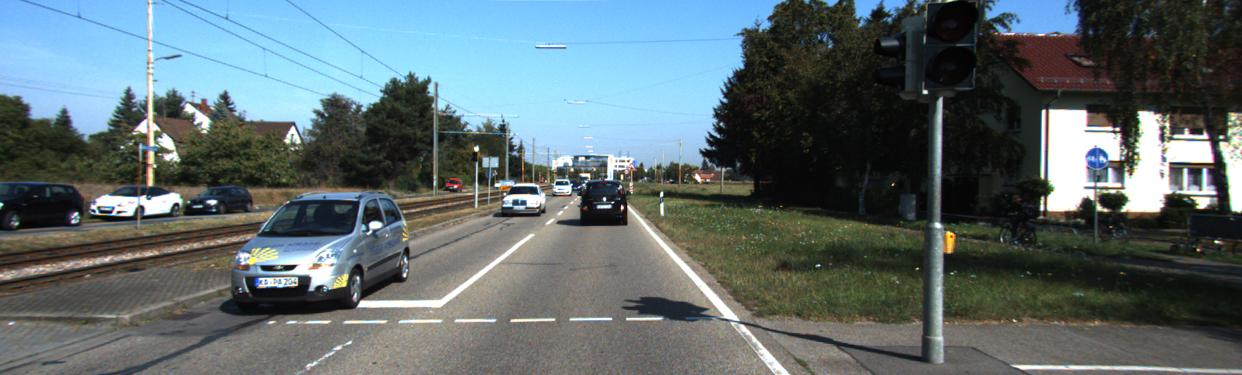} & \includegraphics[width=47mm]{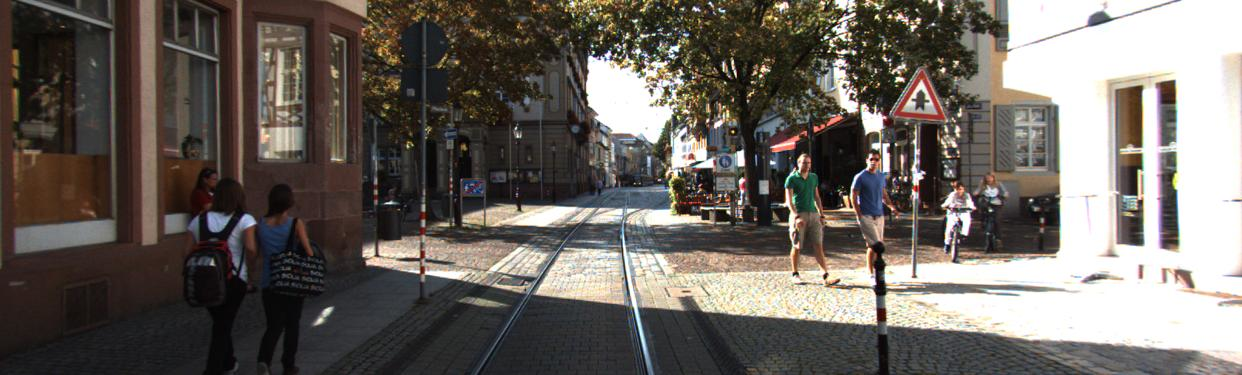}  \\
\begin{sideways}\bf \centering \ \ \ \  GT\end{sideways} & \includegraphics[width=47mm]{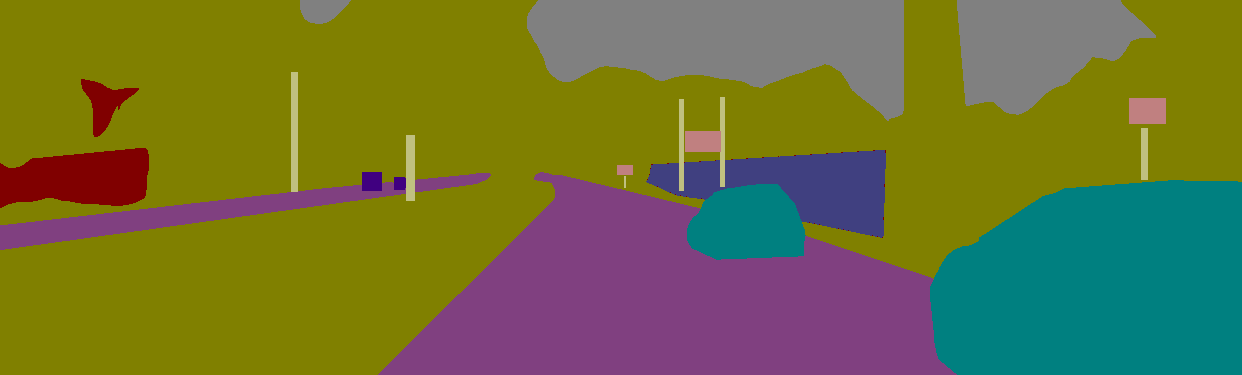} & \includegraphics[width=47mm]{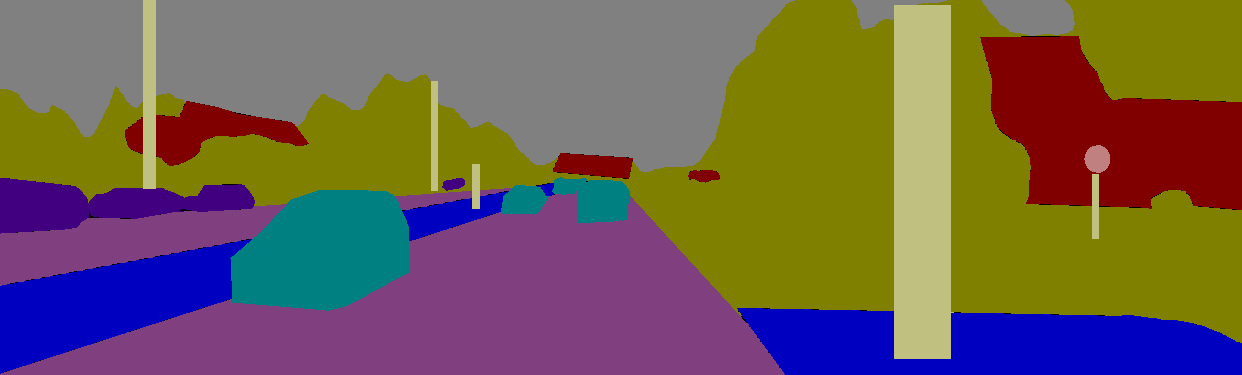} & \includegraphics[width=47mm]{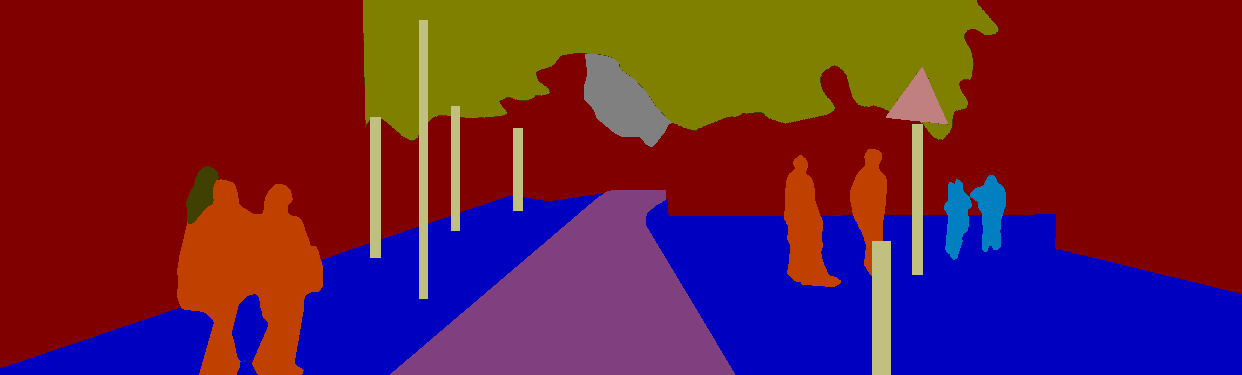}\\
\begin{sideways}\bf\centering Baseline\end{sideways} & \includegraphics[width=47mm]{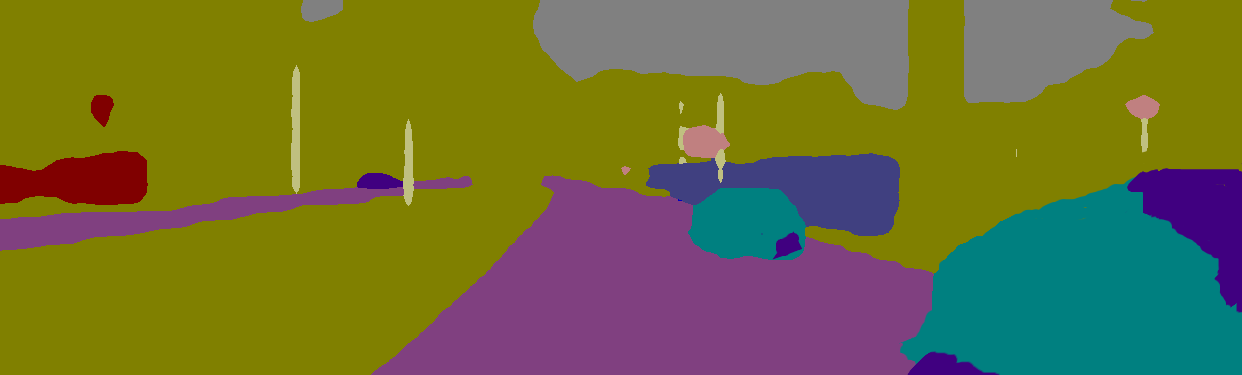} & \includegraphics[width=47mm]{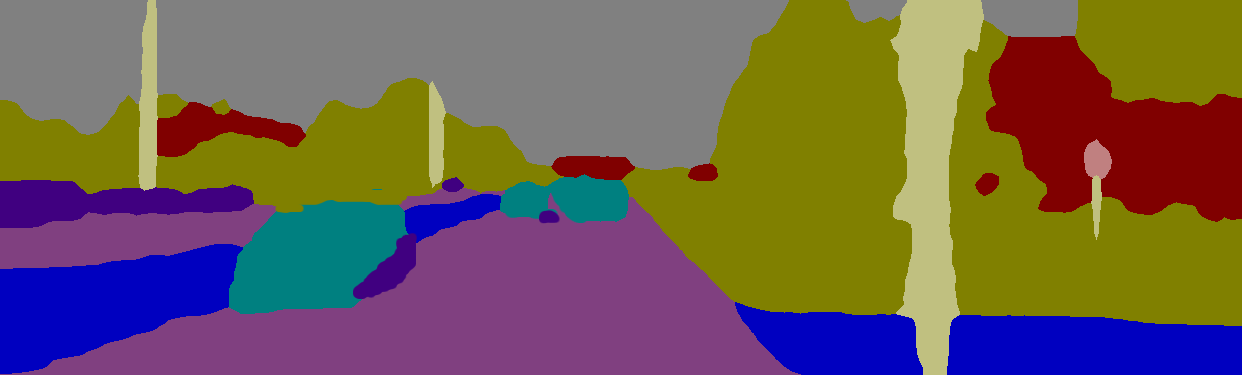} & \includegraphics[width=47mm]{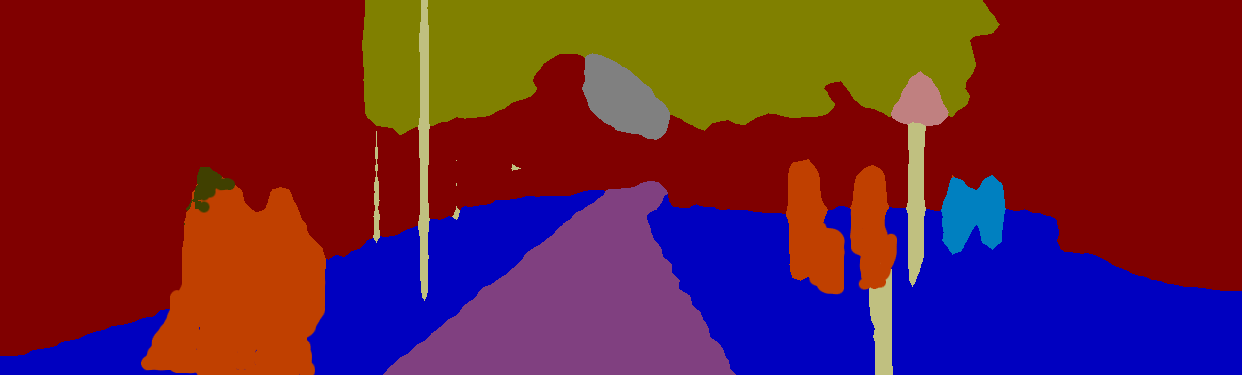}\\
\begin{sideways}\bf \centering \ \ Joint\end{sideways} & \includegraphics[width=47mm]{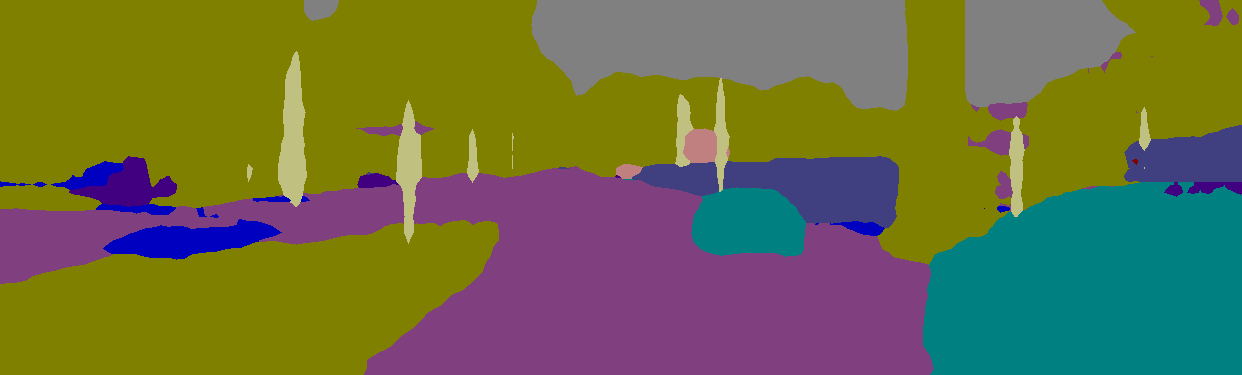} & \includegraphics[width=47mm]{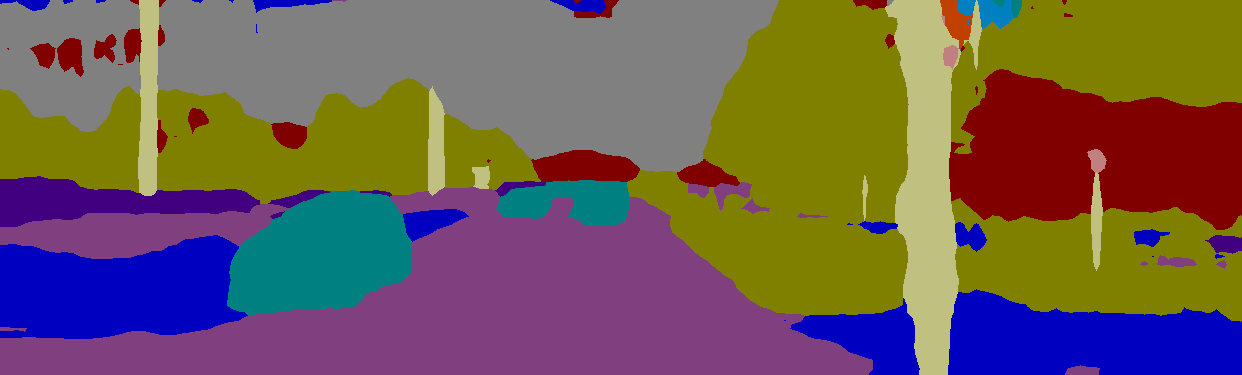} & \includegraphics[width=47mm]{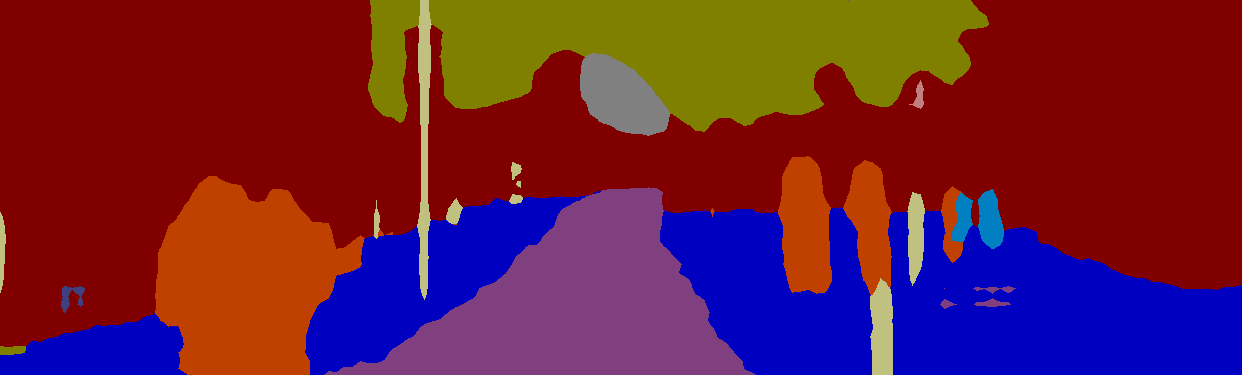} \\
\begin{sideways}\bf \centering +Context\end{sideways} & \includegraphics[width=47mm]{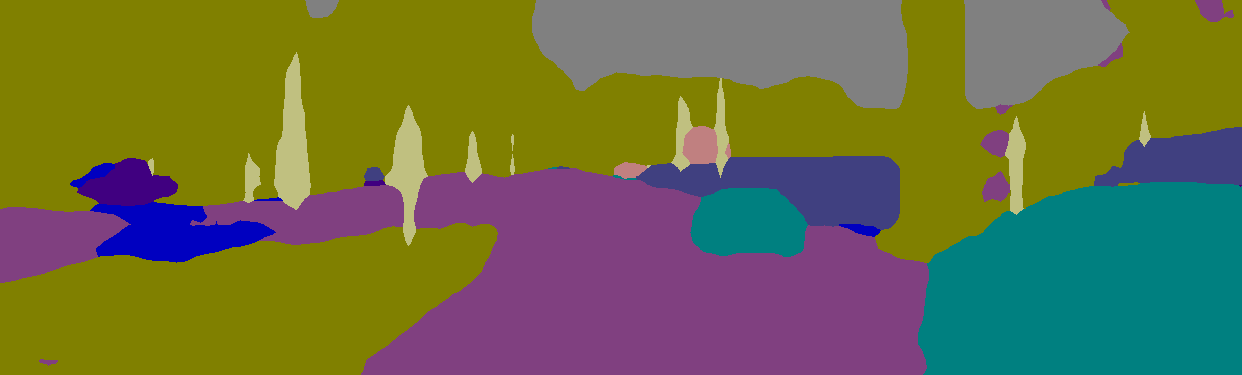} & \includegraphics[width=47mm]{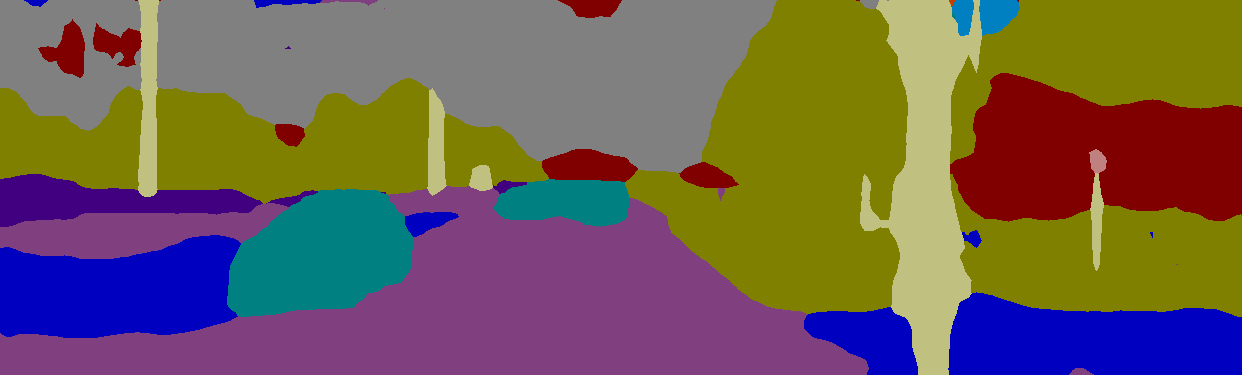} & \includegraphics[width=47mm]{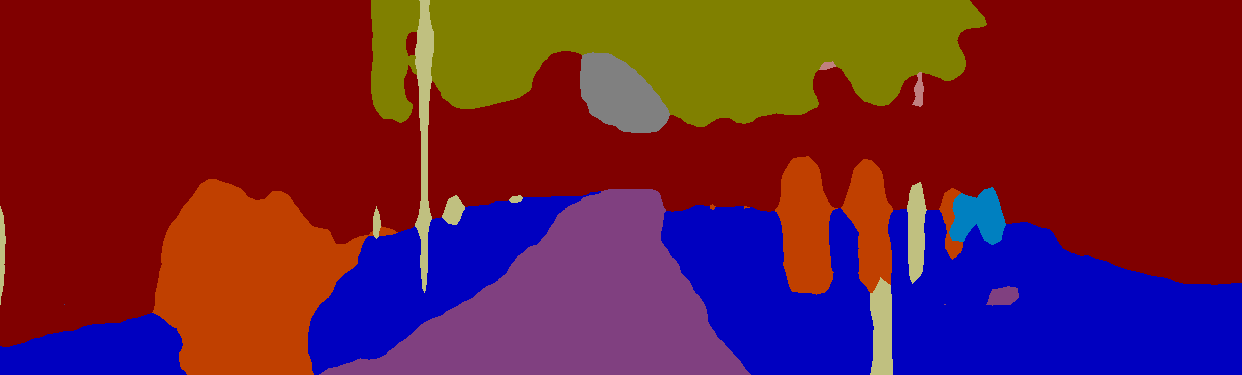}\\

& \multicolumn{3}{c}{\includegraphics[width=148mm, height=8mm]{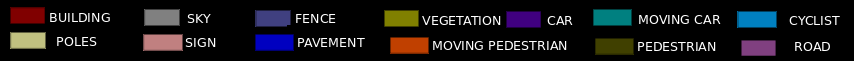}}\\
 
\end{tabular}

\end{center}

 \caption{Qualitative evaluation of joint labels with Ground Truth annotations on our KITTI test dataset. Top to Bottom: (1) Input image from KITTI sequences (2) Ground Truth for semantic motion segmentation (3) Baseline predictions: joint labels using dilated convolution. (4) Joint Module: Results obtained after feature amplification with optical flow.(5) Context Module: joint predictions after feature enhancement with context module.(Best viewed in color)}
 \label{fig:qualitative1}
 \end{figure*}
\subsection{Results}
We evaluate the proposed approach on our manually annotated KITTI Tracking test dataset. The testing images(see Sec. \ref{dataset}) chosen from different sequences pose challenging scenarios for motion segmentation with multiple moving objects. Also, there are prominent cases where moving cars lie in the camera subspace. To demonstrate qualitative results we take four sequences consisting of 116, 143, 309 and 46 images. Qualitative results are provided in Fig. \ref{fig:qualitative1}, Fig. \ref{fig:qualitative2}  and on complete sequences in the \textit{supplementary video}.  To the best of our knowledge, there are no available monocular joint semantic and motion baseline. Hence, we show independent semantic and motion evaluation with the existing state of art in the respective fields. For instance, for a pixel bearing joint label - 'Moving Car', we say 'Car' as the semantic label or object class and 'Moving' as the motion class of the pixel. Comparative evaluations are carried out for semantic segmentation and monocular motion segmentation. However, for joint semantic and motion labels, we demonstrate evaluations against manually annotated Ground Truth labels.     

\subsubsection{Qualitative Evaluation}
In this section, we show our results with joint labels for different stages proposed in the paper, in comparison to Ground Truth. We also show qualitative assessment of motion segmentation in monocular settings with STMOP-M. \cite{fragkiadaki2015learning}. In the Figures, 'Stationary Car' and 'Stationary Pedestrian' labels are abbreviated as 'Car' and 'Pedestrian' respectively, while the label is prefixed with 'Moving' in case of motion.\\\\
\begin{table}
\begin{center}
\begin{tabular}[l]{ |c|c|c| }
 \hline
 \textbf{Model} & \textbf{Stationary} & \textbf{Moving} \\
 \hline
 STMOP-M & 98.34 & 83.91 \\
 \hline  
 SHEAR-M& 99.85 & 84.37 \\
 \hline
 \textbf{Ours} ( Joint+Context ) & 99.55 & \textbf{89.28} \\
 \hline
\end{tabular}
\end{center}
\caption{Quantitative evaluation on our KITTI test tracking dataset. We compare  PPV (Positive predicted value) from our approach  with the state of the art sparse motion segmentation SHEAR-M\cite{tourani2016using} and STMOP-M\cite{fragkiadaki2015learning}. We achieve 4.9\% gain in the metric over the existing state of art.}
\label{table:3}
\end{table}
\textbf{Motion}: We show improvements over our baseline results in Fig. \ref{fig:baselineimprovement}. Baseline results labels parts of moving car as stationary. However, with optical flow based feature amplification, pixels for cars in motion are rectified as moving. Further, via feature enhancement with Context Module, labels improve significantly. We attribute the improvements shown by using feature amplification, to the fact that temporal consistency has been incorporated using optical flow into the baseline. Further, the proposed feature amplification has clear demarcation between the boundaries of the moving objects and stationary surroundings due to variance in flow vector magnitude, which is being incorporated into the final segmentation.

STMOP \cite{fragkiadaki2015learning} generates moving object proposals on video sequences. We use the code available and generate proposals on KITTI sequences. For fair comparison, we take the proposals with best supervoxel projection on the objects. We show our monocular motion segmentation results in comparison to Ground Truth and STMOP\cite{fragkiadaki2015learning} moving object proposals. In Fig. \ref{fig:qualitative3}, consisting of images with single and multiple moving cars, STMOP-M leads to over segmentation, while our approach correctly segments the moving car, also removing extra segments of road and fence. In the above cases, STMOP fails in outdoor robotic scenarios essentially due to large camera motion and optical flow \textit{bleeding}, while our approach uses semantic priors and benefits from motion and semantic correlation.
\begin{table*}[t]

\caption {Quantitative Evaluation - Left: Single moving object, Right: Multiple moving objects}
\begin{tabular}[l]{ |c|c|c| }
 \hline
 \textbf{Model} & \textbf{Stationary} & \textbf{Moving} \\
 \hline
 \cite{fragkiadaki2015learning} & 97.75 & 62.97 \\
 \hline  
 Baseline & \textbf{99.44} & 76.36 \\
 \hline
 Joint & 99.35 & 81.94 \\
 \hline
 Joint+Context & 99.28 & \textbf{83.69} \\
 \hline
\end{tabular}
\hfill
\begin{tabular}[l]{ |c|c|c| }
 \hline
 
 \textbf{Model} & \textbf{Stationary} & \textbf{Moving} \\
 \hline
 \cite{fragkiadaki2015learning} & 97.63 & 44.53 \\
 \hline  
 Baseline & \textbf{99.05} & 66.23 \\
 \hline
 Joint & 99.03 & 70.67 \\
 \hline
 Joint+Context & 98.97 & \textbf{71.98} \\
 \hline
\end{tabular}
\hfill
\\\\


\caption{Quantitative analysis of motion label predictions with STMOP. Left: On our annotated Kitti(tracking sequence 4) test dataset - consisting of lone moving object. Right: On our annotated kitti images, consisting of multiple moving cars.  We compare our results with \cite{fragkiadaki2015learning} moving object proposals. }
\label{table:2}
\end{table*}
\begin{table*}
\begin{center}
\setlength\tabcolsep{5.0pt}
\begin{tabular}[t]{ |c|c|c|c|c|c|c|c|c|c|c|c|c| }
 \hline
Method
 & \rot{Building} & \rot{Vegetation} & \rot{Sky} & \rot{Car} & \rot{Sign} & \rot{Road} & \rot{Pedestrian} & \rot{Fence} & \rot{Pole} & \rot{Sidewalk} & \rot{Cyclist} & \rot{mean IOU} \\
 \hline \Tstrut
 
 Segnet   & 66.70 & 78.11 & 89.32 & 69.74 & 12.45 & 71.69& 12.09 & 25.03 & 21.12 & 44.01& 11.2 & 45.61 \\
 
Deeplab  & 73.35 & 84.17 & 91.33 & 70.76 & 7.66 & 69.63& 24.41 & 68.30 & 16.51 & 26.14 & 13.53 & 49.62 \\

 Ours-S
& 78.52 & 84.99 & 90.07 & 88.18& 19.28 & 75.82 & 8.46 & 76.60 & 29.31 & 36.84 & 66.70& \textbf{59.53} \\
 \hline
\end{tabular}
\end{center}
\caption{Quantitative evaluation of semantic label predictions from our proposed approach - Joint+Context (Ours - S) on our KITTI test dataset. We compare our method with DeepLab-LFOV\cite{chen2014semantic} and Segnet\cite{badrinarayanan2015segnet2}, known for semantic segmentation on outdoor driving scenes.}
\label{table:1}
\end{table*}
\\
\textbf{Joint Semantic and Motion}: We also evaluate our approach with the Ground Truth semantic motion labeling. In the sequences demonstrated in Fig. \ref{fig:qualitative1}, the Baseline results incorrectly labels patches of the moving car closer to the camera(in Sequence 1) as stationary(seen with \textit{Violet} color). Similar observation is found in the  baseline results of the moving car in Sequence 2. The patches are rectified as moving as a consequence of joint training with amplified features. This again reiterates the utility of joint learning and inference between motion and semantic cues. The improper patches on the moving cars in the sequences are further rectified by the context module using multi scale context aggregation. To perceive joint segmentation, other than car scenes, we consider a sequence(Sequence 3) from KITTI with Moving Pedestrians. The results for each stage are depicted in Fig. \ref{fig:qualitative1}. Parts of moving pedestrians on the left are labeled as stationary in baseline results. The joint learning with context aggregation corrects the motion domain of pedestrians. Also, for pedestrians far away from the camera, false positive cases are observed from our approach in tiny patches due to inconsistency in optical flow magnitude of the pedestrian with large distance from the camera. Further, for consistency of our joint labels in challenging outdoor scenes, we show joint semantic motion results on both highway and city street scenes in Fig. \ref{fig:qualitative3}.

\subsubsection{Quantitative Evaluation}
In this section, we perform a quantitative assessment of both semantic and motion segmentation. We show evaluations with \cite{tourani2016using} and \cite{fragkiadaki2015learning}. For semantic segmentation we compare our results with \cite{chen14semantic} and \cite{badrinarayanan2015segnet2}, which have shown results for semantic segmentation on driving scenes.\\\\
\textbf{Motion}: For quantitative evaluation of motion segmentation, we compare our results with STMOP moving object proposals. Evaluation is staged by cross verification of each predicted pixel with corresponding ground truth motion label - stationary or moving. The evaluation is unfolded in two models. First, we compare our dense motion segmentation with STMOP moving object proposal. We use intersection over union as the evaluation metric for dense motion segmentation. The metric is defined as \textit{TP/(TP+FP+FN)}, where TP denotes true positive, FP false positive and FN false negative. Table \ref{table:2} summarizes our quantitative motion segmentation evaluation. The assessment is done in two broad categories,i.e, on annotated sequences with lone moving object and sequences with multiple objects in motion. In the case with single moving car, we achieved 70.67\% accuracy in detection of the moving car from our joint module, while STMOP yields 59.97\% detection accuracy. The increase in accuracy is attributed to incorporated label and motion correlation. Further, using context aggregation, the context module yields further improvement in the efficiency. In case of multiple moving objects, STMOP yields 41.53\% efficiency. The decrease in accuracy from STMOP is due to large camera motion observed in the scenes, while our joint module provides 70.67\% success rate. The joint learning exploits the fact that the likelihood of a moving tree or pole is less compared to a moving car or moving person, resulting in substantial improvement in motion segmentation. Another keynote observation would be a slight decrease in stationary accuracy over our baseline results. This is due to the fact that different objects can exhibit different optical flow depending on the depth from the camera, even though they share the same global motion. The decrease is although marginal as shown in Table \ref{table:2}. We also show motion segmentation evaluation with existing state of art in sparse monocular motion segmentation. The IOU metric used for dense motion segmentation is not known to be used in case of sparse evaluations. Therefore, for fair comparison with sparse segmentation, we use positive predictive value (PPV) or precision- (TP/TP+FP) as the evaluation metric. The results are summarized in Table \ref{table:3}. We gain 4.9\% in motion label precision over the state of art SHEAR-M\cite{tourani2016using} on our test dataset.
\begin{figure*}[t!]
\setlength\tabcolsep{2.0pt}
\begin{center}
\begin{tabular}{c c c}

\includegraphics[width=50mm]{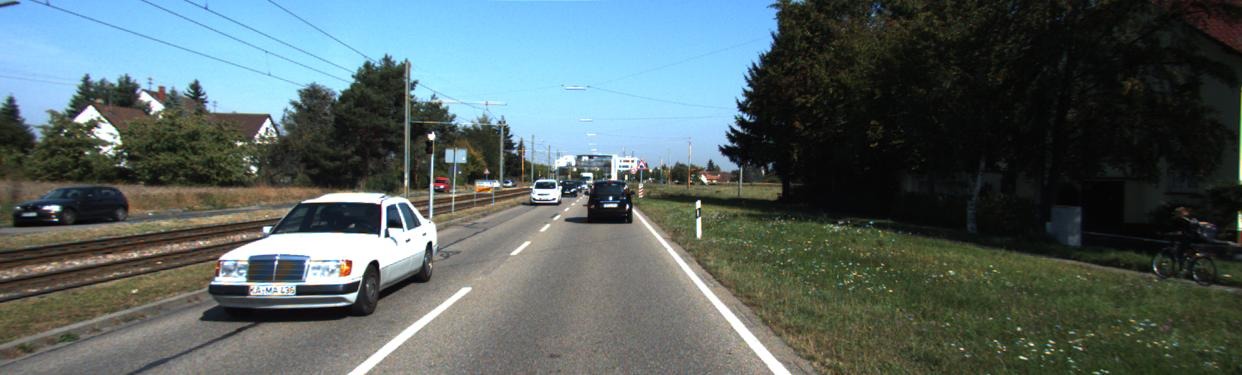} & \includegraphics[width=50mm]{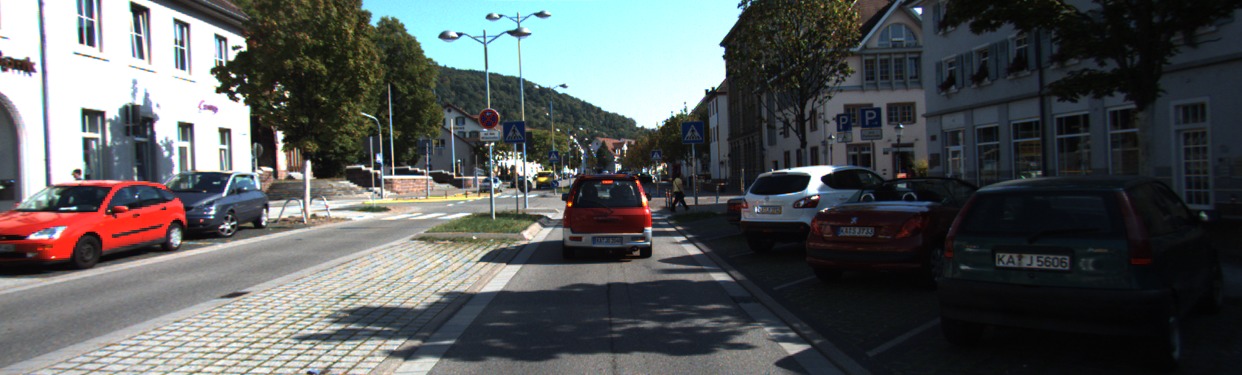} &
\includegraphics[width=50mm]{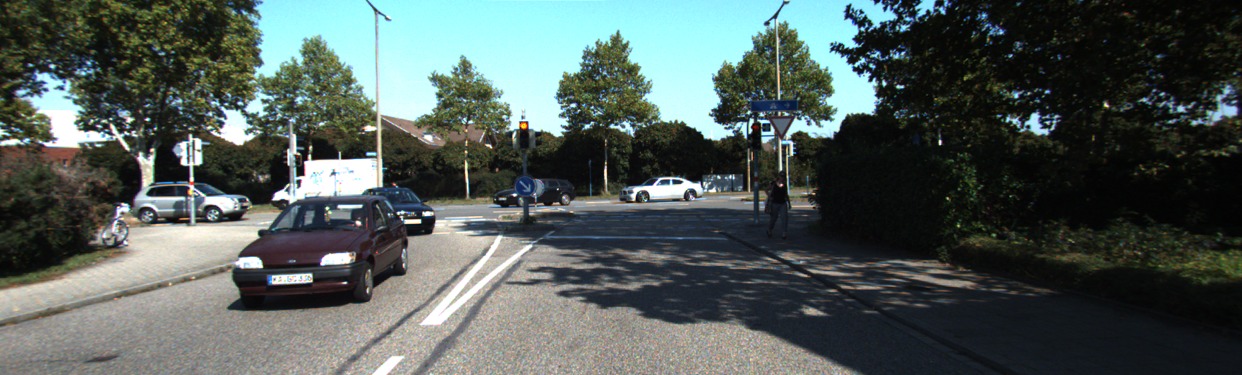} \\ \includegraphics[width=50mm]{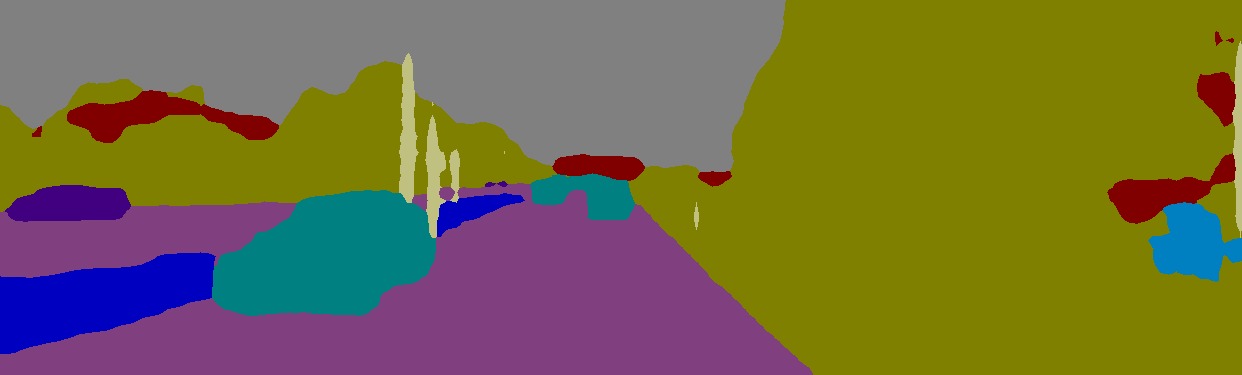} &
\includegraphics[width=50mm]{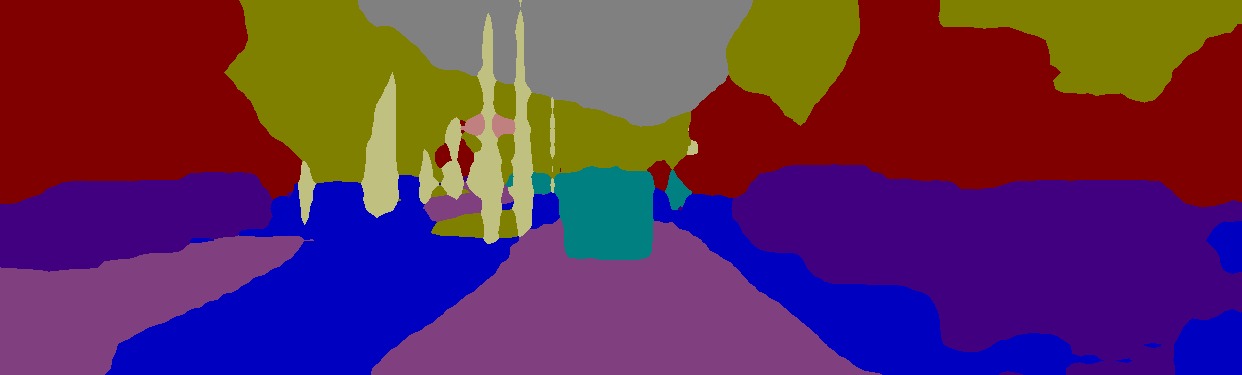} & \includegraphics[width=50mm]{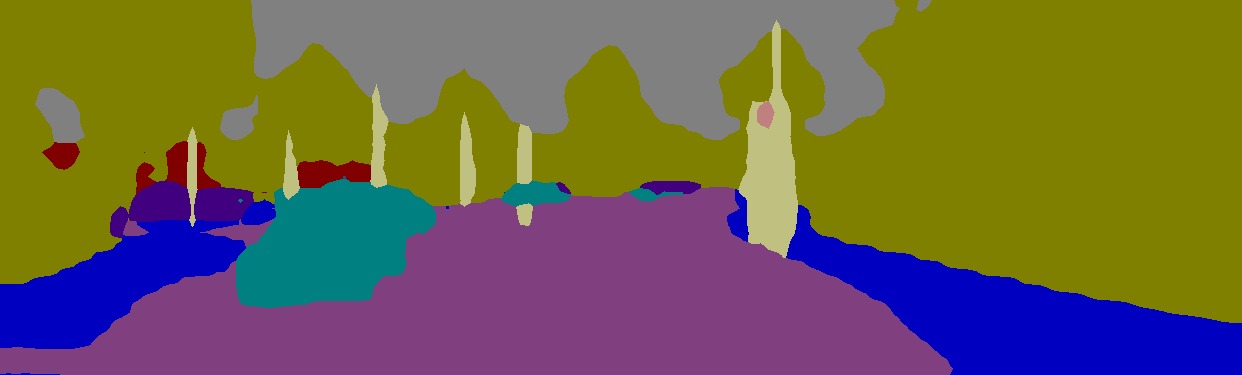} \\

\multicolumn{3}{c}{\includegraphics[width=153mm, height=9mm]{BaselineImprovementFinal/palette_big}}\\
\end{tabular}

\end{center}

 \caption{Joint Semantic and motion labeling obtained from the proposed approach on challenging urban scenes. Specifically, in the figure, from Left to Right: Highway scene, City Streets and a drive scene with relatively less traffic. The joint labels obtained in these settings depict robustness and consistency of our proposed approach.(Best viewed in color)}
 \label{fig:qualitative2}
 \end{figure*}

\textbf{Semantics:}
For quantitative evaluation of semantic image segmentation, we use per class Intersection over Union similar to the metric used for dense motion segmentation evaluation. This is done for 11 semantic labels on our KITTI test dataset. We perform quantitative semantic evaluation of our approach against Segnet\cite{badrinarayanan2015segnet2}, which has shown results on outdoor driving scenes such as KITTI, and DeepLab-LFOV \cite{chen14semantic}. For comparison with  \cite{chen14semantic} we use the publicly available pre-trained model on PASCAL dataset and fine tune it on our KITTI training dataset. We run both the algorithms, Segnet and DeepLab, on our KITTI test dataset. The \textit{semantic label} accuracy of the models on the test set is reported in Table \ref{table:1}. Our approach (Joint+Context) outperforms the other two architectures. This is due to the fact that dilated architecture produces higher resolution output crucial to dense prediction in comparison to the strided and pooled architectures in the former propositions.                         

 
\section{CONCLUSIONS}
In this paper, we have proposed a joint approach to predict semantic and motion labels using a monocular camera. We incorporate spatial and temporal information to learn object class and motion labels jointly. Evaluations show an increase in pixel wise motion segmentation accuracy without using stereo information. We learn pixel wise labels without the need for training temporal networks for motion cues, which has proved to be a pitfall with unavailability of large annotated datasets. To contribute and encourage future works on monocular semantic motion segmentation, we plan to release the annotated dataset and trained models.

\section{ACKNOWLEDGEMENTS}
We would like to thank J. Krishna Murthy for proofreading and paper editing. We are also grateful to Parv Parkhiya and Aman Bansal for help with dataset annotation on KITTI Tracking benchmark.




\bibliographystyle{apalike}
{\small
\bibliography{example}}

\begin{thebibliography}{}

\bibitem[Athanasiadis et~al., 2007]{athanasiadis2007semantic}
Athanasiadis, T., Mylonas, P., Avrithis, Y., and Kollias, S. (2007).
\newblock Semantic image segmentation and object labeling.
\newblock {\em IEEE transactions on circuits and systems for video technology},
  17(3):298--312.

\bibitem[Badrinarayanan et~al., 2015]{badrinarayanan2015segnet2}
Badrinarayanan, V., Handa, A., and Cipolla, R. (2015).
\newblock Segnet: A deep convolutional encoder-decoder architecture for robust
  semantic pixel-wise labelling.
\newblock {\em arXiv preprint arXiv:1505.07293}.

\bibitem[Chen et~al., 2014]{chen2014semantic}
Chen, L.-C., Papandreou, G., Kokkinos, I., Murphy, K., and Yuille, A.~L.
  (2014).
\newblock Semantic image segmentation with deep convolutional nets and fully
  connected crfs.
\newblock {\em arXiv preprint arXiv:1412.7062}.

\bibitem[Chen et~al., 2015]{chen14semantic}
Chen, L.-C., Papandreou, G., Kokkinos, I., Murphy, K., and Yuille, A.~L.
  (2015).
\newblock Semantic image segmentation with deep convolutional nets and fully
  connected crfs.
\newblock In {\em ICLR}.

\bibitem[Dai et~al., 2015]{dai2015boxsup}
Dai, J., He, K., and Sun, J. (2015).
\newblock Boxsup: Exploiting bounding boxes to supervise convolutional networks
  for semantic segmentation.
\newblock In {\em Proceedings of the IEEE International Conference on Computer
  Vision}, pages 1635--1643.

\bibitem[Elhamifar and Vidal, 2009]{elhamifar2009sparse}
Elhamifar, E. and Vidal, R. (2009).
\newblock Sparse subspace clustering.
\newblock In {\em Computer Vision and Pattern Recognition, 2009. CVPR 2009.
  IEEE Conference on}, pages 2790--2797. IEEE.

\bibitem[Fields, 2001]{fields2001probabilistic}
Fields, R. (2001).
\newblock Probabilistic models for segmenting and labeling sequence data.
\newblock In {\em ICML 2001}.

\bibitem[Fischer et~al., 2015]{fischer2015flownet}
Fischer, P., Dosovitskiy, A., Ilg, E., H{\"a}usser, P., Haz{\i}rba{\c{s}}, C.,
  Golkov, V., van~der Smagt, P., Cremers, D., and Brox, T. (2015).
\newblock Flownet: Learning optical flow with convolutional networks.
\newblock {\em arXiv preprint arXiv:1504.06852}.

\bibitem[Fragkiadaki et~al., 2015]{fragkiadaki2015learning}
Fragkiadaki, K., Arbel{\'a}ez, P., Felsen, P., and Malik, J. (2015).
\newblock Learning to segment moving objects in videos.
\newblock In {\em 2015 IEEE Conference on Computer Vision and Pattern
  Recognition (CVPR)}, pages 4083--4090. IEEE.

\bibitem[Geiger et~al., 2012]{Geiger2012CVPR}
Geiger, A., Lenz, P., and Urtasun, R. (2012).
\newblock Are we ready for autonomous driving? the kitti vision benchmark
  suite.
\newblock In {\em Conference on Computer Vision and Pattern Recognition
  (CVPR)}.

\bibitem[Glorot and Bengio, 2010]{glorot2010understanding}
Glorot, X. and Bengio, Y. (2010).
\newblock Understanding the difficulty of training deep feedforward neural
  networks.
\newblock In {\em Aistats}, volume~9, pages 249--256.

\bibitem[Jia et~al., 2014]{jia2014caffe}
Jia, Y., Shelhamer, E., Donahue, J., Karayev, S., Long, J., Girshick, R.,
  Guadarrama, S., and Darrell, T. (2014).
\newblock Caffe: Convolutional architecture for fast feature embedding.
\newblock {\em arXiv preprint arXiv:1408.5093}.

\bibitem[Karpathy et~al., 2014]{karpathy2014large}
Karpathy, A., Toderici, G., Shetty, S., Leung, T., Sukthankar, R., and Fei-Fei,
  L. (2014).
\newblock Large-scale video classification with convolutional neural networks.
\newblock In {\em Proceedings of the IEEE conference on Computer Vision and
  Pattern Recognition}, pages 1725--1732.

\bibitem[Koltun, 2011]{koltun2011efficient}
Koltun, V. (2011).
\newblock Efficient inference in fully connected crfs with gaussian edge
  potentials.
\newblock {\em Adv. Neural Inf. Process. Syst}.

\bibitem[LeCun et~al., 1989]{lecun1989backpropagation}
LeCun, Y., Boser, B., Denker, J.~S., Henderson, D., Howard, R.~E., Hubbard, W.,
  and Jackel, L.~D. (1989).
\newblock Backpropagation applied to handwritten zip code recognition.
\newblock {\em Neural computation}, 1(4):541--551.

\bibitem[Lin et~al., 2015]{lin2015efficient}
Lin, G., Shen, C., Reid, I., et~al. (2015).
\newblock Efficient piecewise training of deep structured models for semantic
  segmentation.
\newblock {\em arXiv preprint arXiv:1504.01013}.

\bibitem[Liu et~al., 2015]{liu2015semantic}
Liu, Z., Li, X., Luo, P., Loy, C.-C., and Tang, X. (2015).
\newblock Semantic image segmentation via deep parsing network.
\newblock In {\em Proceedings of the IEEE International Conference on Computer
  Vision}, pages 1377--1385.

\bibitem[Long et~al., 2015]{long2015fully}
Long, J., Shelhamer, E., and Darrell, T. (2015).
\newblock Fully convolutional networks for semantic segmentation.
\newblock In {\em Proceedings of the IEEE Conference on Computer Vision and
  Pattern Recognition}, pages 3431--3440.

\bibitem[Park et~al., 2016]{park2016combining}
Park, E., Han, X., Berg, T.~L., and Berg, A.~C. (2016).
\newblock Combining multiple sources of knowledge in deep cnns for action
  recognition.
\newblock In {\em 2016 IEEE Winter Conference on Applications of Computer
  Vision (WACV)}, pages 1--8. IEEE.

\bibitem[Reddy et~al., 2014]{reddy2014semantic}
Reddy, N.~D., Singhal, P., and Krishna, K.~M. (2014).
\newblock Semantic motion segmentation using dense crf formulation.
\newblock In {\em Proceedings of the 2014 Indian Conference on Computer Vision
  Graphics and Image Processing}, page~56. ACM.

\bibitem[Rozantsev et~al., 2014]{rozantsev2014flying}
Rozantsev, A., Lepetit, V., and Fua, P. (2014).
\newblock Flying objects detection from a single moving camera.
\newblock {\em arXiv preprint arXiv:1411.7715}.

\bibitem[Russell et~al., 2009]{russell2009associative}
Russell, C., Kohli, P., Torr, P.~H., et~al. (2009).
\newblock Associative hierarchical crfs for object class image segmentation.
\newblock In {\em 2009 IEEE 12th International Conference on Computer Vision},
  pages 739--746. IEEE.

\bibitem[Shotton et~al., 2008]{shotton2008semantic}
Shotton, J., Johnson, M., and Cipolla, R. (2008).
\newblock Semantic texton forests for image categorization and segmentation.
\newblock In {\em Computer vision and pattern recognition, 2008. CVPR 2008.
  IEEE Conference on}, pages 1--8. IEEE.

\bibitem[Simonyan and Zisserman, 2014a]{simonyan2014two}
Simonyan, K. and Zisserman, A. (2014a).
\newblock Two-stream convolutional networks for action recognition in videos.
\newblock In {\em Advances in Neural Information Processing Systems}, pages
  568--576.

\bibitem[Simonyan and Zisserman, 2014b]{simonyan2014very}
Simonyan, K. and Zisserman, A. (2014b).
\newblock Very deep convolutional networks for large-scale image recognition.
\newblock {\em arXiv preprint arXiv:1409.1556}.

\bibitem[Tokmakov et~al., 2016]{tokmakov2016weakly}
Tokmakov, P., Alahari, K., and Schmid, C. (2016).
\newblock Weakly-supervised semantic segmentation using motion cues.
\newblock {\em arXiv preprint arXiv:1603.07188}.

\bibitem[Tourani and Krishna, 2016]{tourani2016using}
Tourani, S. and Krishna, K.~M. (2016).
\newblock Using in-frame shear constraints for monocular motion segmentation of
  rigid bodies.
\newblock {\em Journal of Intelligent \& Robotic Systems}, 82(2):237--255.

\bibitem[Wedel et~al., 2009]{wedel2009detection}
Wedel, A., Mei{\ss}ner, A., Rabe, C., Franke, U., and Cremers, D. (2009).
\newblock Detection and segmentation of independently moving objects from dense
  scene flow.
\newblock In {\em International Workshop on Energy Minimization Methods in
  Computer Vision and Pattern Recognition}, pages 14--27. Springer.

\bibitem[Weinzaepfel et~al., 2013]{weinzaepfel2013deepflow}
Weinzaepfel, P., Revaud, J., Harchaoui, Z., and Schmid, C. (2013).
\newblock Deepflow: Large displacement optical flow with deep matching.
\newblock In {\em Proceedings of the IEEE International Conference on Computer
  Vision}, pages 1385--1392.

\bibitem[Yu and Koltun, 2015]{yu2015multi}
Yu, F. and Koltun, V. (2015).
\newblock Multi-scale context aggregation by dilated convolutions.
\newblock {\em arXiv preprint arXiv:1511.07122}.

\end{thebibliography}

\vfill
\end{document}